\PassOptionsToPackage{table}{xcolor}
\PassOptionsToPackage{capitalize,noabbrev,nameinlink}{cleveref}
\documentclass{fairmeta}
\usepackage[utf8]{inputenc}
\usepackage{latexsym}
\usepackage{inconsolata}
\usepackage{amsmath,amssymb,mathtools}
\usepackage{array,float}
\usepackage[normalem]{ulem}
\usepackage{xurl}
\usepackage{wrapfig,needspace}
\makeatletter
\newcommand{\FinishWrap}{\par
  \ifnum\c@WF@wrappedlines>1
    \dimen@=\baselineskip
    \multiply\dimen@ by \numexpr\c@WF@wrappedlines-1\relax
    \vskip\dimen@
  \fi
  \WFclear}
\makeatother
\newlength{\ArxivColumnWidth}
\title{LongR: Unleashing Long-Context Reasoning via Reinforcement Learning with Dense Utility Rewards}
\author[1,\dagger]{Qize~Yu}
\author[2,\dagger]{Zijun~Chen}
\author[1,\dagger]{Chen~Wu}
\author[1]{Yue~Chen}
\author[3]{Yiyao~Yu}
\author[4]{Tingfeng~Hui}
\author[1,*,\ddagger]{Bowen~Ping}
\author[2]{Junchi~Yan}
\author[1,*]{Baobao~Chang}
\affiliation[1]{Peking University}
\affiliation[2]{Shanghai Jiao Tong University}
\affiliation[3]{Tsinghua University}
\affiliation[4]{Beijing University of Posts and Telecommunications}
\contribution[\dagger]{Equal contribution.\quad${}^{*}$Corresponding authors.\quad${}^{\ddagger}$Project leader.}
\date{October 12, 2026}
\abstract{Reinforcement Learning (RL) has emerged as a key driver for LLM reasoning. This capability is equally pivotal in long-context scenarios---such as long-dialogue understanding and structured data analysis---where the challenge extends beyond consuming tokens to performing rigorous deduction. While existing efforts focus on data synthesis or architectural changes, recent work points out that relying solely on sparse, outcome-only rewards yields limited gains, as such coarse signals are often insufficient to effectively guide the complex long-context reasoning. To address this, we propose \textbf{LongR}, a unified framework that enhances long-context performance by integrating a dynamic ``Think-and-Read'' mechanism---which interleaves reasoning with document consultation---with a contextual density reward based on relative information gain to quantify the utility of the relevant documents, without evidence annotations. Empirically, LongR achieves a gain of over 9\% on LongBench v2 and consistent improvements on RULER and InfiniteBench, demonstrating robust efficiency in navigating extensive contexts. Furthermore, LongR consistently enhances performance across diverse RL algorithms (e.g., DAPO, GSPO). Finally, we conduct in-depth analyses to investigate the impact of reasoning chain length on efficiency and the model's robustness against distractors.
}
\hypersetup{
pdftitle={LongR: Unleashing Long-Context Reasoning via Reinforcement Learning with Dense Utility Rewards},
pdfauthor={Qize Yu, Zijun Chen, Chen Wu, Yue Chen, Yiyao Yu, Tingfeng Hui, Bowen Ping, Junchi Yan, Baobao Chang}
}
\begin{document}
\maketitle
\section{Introduction}

Reinforcement Learning (RL) has emerged as a potent mechanism for unlocking the reasoning potential of Large Language Models (LLMs)~\citep{guo2025deepseek,team2025kimi}. However, reasoning is not only critical in domains like mathematics but is equally pivotal in the long-context domain. In real-world scenarios---ranging from synthesizing insights across heterogeneous documents to debugging repository-scale codebases---the challenge lies not merely in consuming vast input tokens, but in performing rigorous deduction and comprehension over the content. Moreover, distinct long-context reasoning capabilities are foundational for autonomous agents to sustain coherent actions over extended trajectories.

\begin{wrapfigure}{R}{\ArxivColumnWidth}

    \centering
    \includegraphics[width=1.0\linewidth]{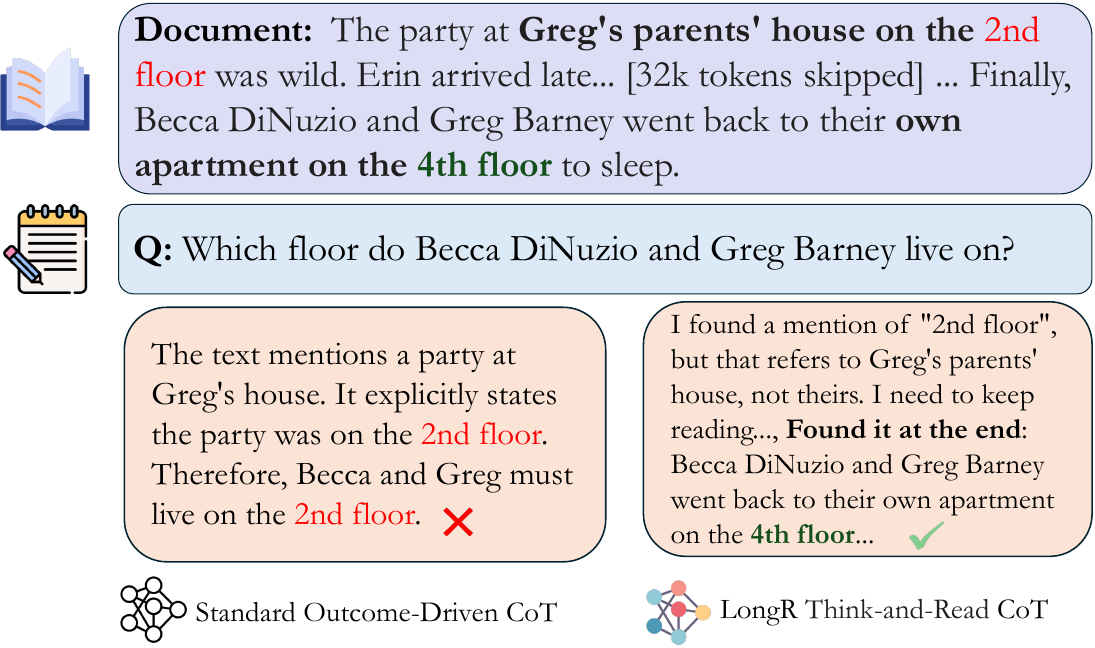}
    \caption{Qualitative comparison between Outcome-Driven CoT (Standard RL, left) and Think-and-Read CoT (LongR, right). Left: Without dense supervision, the baseline model tends to rely on superficial query matching while overlooking detailed context reading. Right: LongR incorporates a contextual dense reward. By incentivizing the model to actively consult the context during the reasoning process, it significantly enhances long-context understanding capabilities.}
    \label{fig:1}
\end{wrapfigure}

Recently, there has been a surge of interest in applying RL to long-context scenarios~\citep{bai2025longbench,zeng2025glm,ping2026longactharnessingintrinsicactivation}. Prior works have predominantly focused on synthesizing high-quality reasoning data~\citep{wang2025loongrl} or modifying model architectures to natively handle extended sequences~\citep{team2025kimi_linear}. More recently, LongRLVR~\citep{longrlvr2025} pointed out that relying solely on sparse rewards yields limited performance gains for long-context RL. They proposed partitioning the context into discrete segments and employing an interleaved mechanism of reasoning and chunk selection to enhance long-context understanding.

However, existing exploration like LongRLVR~\citep{longrlvr2025} requires pre-segmenting documents into discrete chunks and training the policy to output specific chunk identifiers. While effective in specific settings, this paradigm imposes structural constraints that limit broader applicability: (1) \textbf{Information Integrity Risks}: Such rigid segmentation introduces an inherent risk of disrupting data integrity, as guaranteeing that information remains completely uncompromised is non-trivial. Furthermore, this approach necessitates artificial data construction with additional overhead, while the resulting specialized format inherently diverges from mainstream post-training distributions and natural user queries. (2) \textbf{Interrupted Reasoning Flow}: Decoupling the action of ``looking up'' from the flow of ``reasoning'' forces the model to adhere to a specialized output format (e.g., discrete tag generation) rather than natural language reasoning. In contrast, standard LLMs benefit from a coherent chain of thought that mirrors natural language reasoning.

To bridge this gap, we propose \textbf{LongR}, a unified framework operating strictly within standard autoregressive generation. We highlight three key contributions: (1) \textbf{Dynamic ``Think-and-Read'' Mechanism}: This mechanism mirrors the human cognitive process of interleaving reasoning with document consultation by integrating evidence seeking directly into the chain of thought. By eliminating the need for discrete chunking, it preserves the integrity of the original information. (2) \textbf{Contextual Density Reward via Relative Information Gain}: Recognizing that not all contexts readily support the construction of explicitly verifiable rewards (e.g., raw documents lacking fine-grained annotations), we instead quantify the information-theoretic utility of retrieved documents to provide dense supervision. By introducing this intrinsic density reward, we incentivize the model to actively seek high-value information, thereby enhancing its efficiency in utilizing long contexts. (3) \textbf{Self-Learned Reasoning Pattern}: We show that the ``Think-and-Read'' pattern can be acquired from generic, readily available data, without ever constructing long-context-specific ``consult-then-reason'' demonstrations. Through a progressive curriculum, models learn to dynamically interleave consultation with reasoning purely via RL, eliminating the need for costly, long-context-specific data engineering.

We empirically validate LongR using LongBench v2~\citep{bai2025longbench}, where it achieves a performance gain of approximately 9\% shown in~\cref{tb:main}. LongR exhibits remarkable universality, enhancing generalization on generic long-context tasks (e.g., RULER and InfiniteBench~\citep{hsieh2024ruler,zhang-etal-2024-bench}) shown in~\cref{tb:ruler} while consistently boosting performance across a diverse spectrum of RL algorithms, including DAPO, GSPO, CISPO~\citep{yu2025dapo,zheng2025group,minimax2025minimaxm1scalingtesttimecompute} shown in~\cref{tb:algo}. Furthermore, we validate the effectiveness of our proposed contextual reward in~\cref{tb:reward}, where it significantly outperforms outcome-only baselines by approximately 4\%. Additionally,~\cref{tb:curriculum} demonstrates that our curriculum learning strategy contributes substantially to performance gains. Finally, we conduct in-depth analyses on the length of reasoning chains in terms of computational efficiency, as well as LongR's robustness against distractors.

Our contributions can be summarized as follows:
\begin{itemize}
    \item We propose \textbf{LongR}, which leverages an additional contextual reward, computed without annotations, to guide long-context reasoning.
    \item We conduct extensive experiments (using LongBench v2, RULER, etc.) to validate the effectiveness of LongR. For instance, LongR achieves an improvement of over 9\% on LongBench v2.
    \item We conduct in-depth analyses to investigate the efficiency and robustness of LongR.
\end{itemize}

\section{Related Work}

\paragraph{Reinforcement Learning in Long-context Scenario.}
While Reinforcement Learning (RL) has become a staple for enhancing mathematical reasoning~\citep{yu2025dapo,zheng2025group}, its adaptation to long-context domains is a burgeoning field. Prevailing strategies predominantly focus on architectural innovations---such as linear or sparse attention mechanisms~\citep{team2025kimi_linear,team2025minicpm4,gao2025seerattention}---which often necessitate extensive pre-training. Others, \citet{wang2025loongrl} and \citet{shen2025qwenlongl15posttrainingrecipelongcontext} prioritize the synthesis of superior reasoning data. LongRLVR~\citep{longrlvr2025} demonstrates that relying solely on outcome-based rewards yields limited gains for long-context reasoning and supplements sparse outcome signals using pre-segmented chunks. In contrast, our method avoids rigid segmentation, thereby preserving data integrity and ensuring broader applicability. A concurrent work, EAPO~\citep{guan2026evidence}, addresses reward sparsity by training a learned black-box proxy to score evidence quality. However, such learned rewards are prone to instability, necessitating a computationally expensive reward-policy co-evolution loop to refine the reward model during training. In contrast, LongR employs Relative Information Gain, a white-box, information-theoretic metric derived directly from a frozen verifier.

\begin{figure}[tbp]
    \centering
    \includegraphics[width=0.95\linewidth]{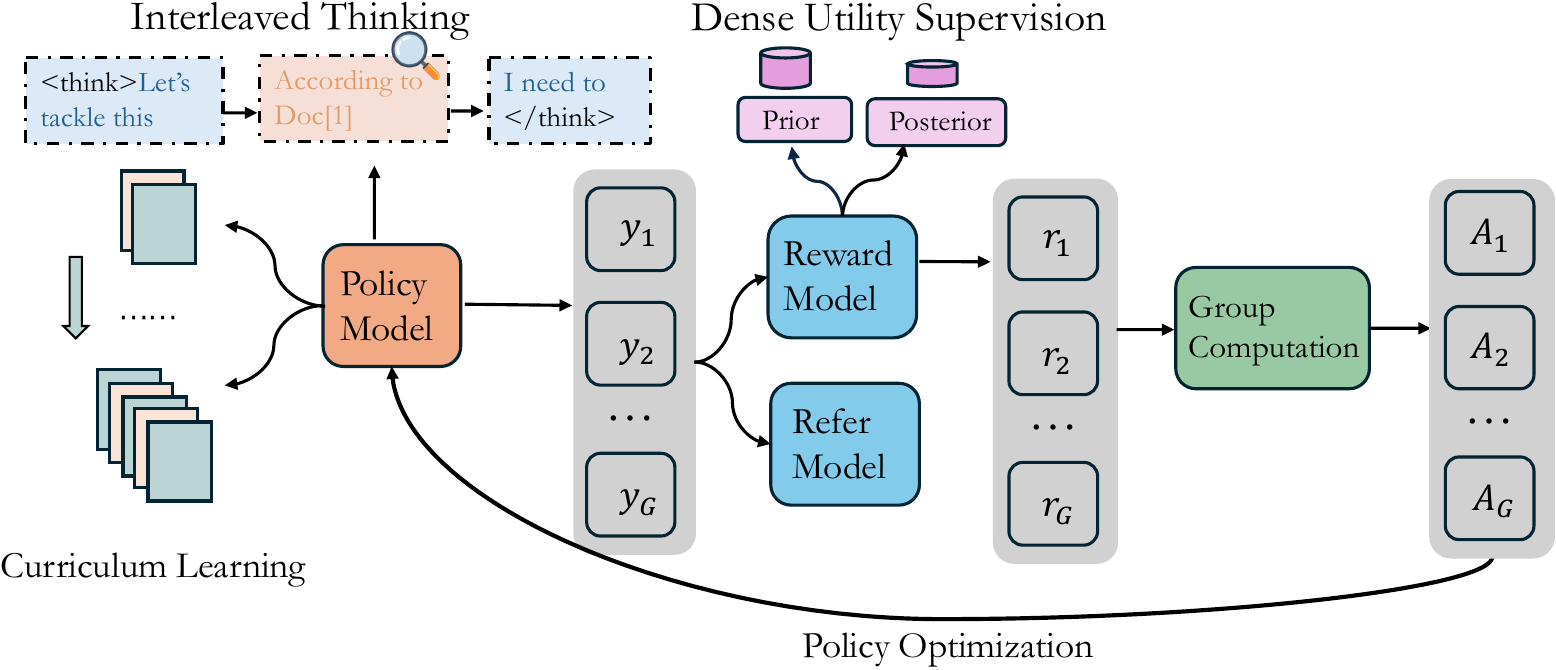}
    \caption{Overview of the LongR Framework. The system operates within a standard reinforcement learning pipeline enhanced by three key mechanisms: (1) Curriculum Learning: through a progressive curriculum, models learn to dynamically interleave consultation with reasoning purely via RL, eliminating the need for costly, long-context-specific SFT data engineering. (2) Interleaved Think-and-Read Policy: generating actions as a dynamic stream where reasoning steps (blue) alternate with grounded evidence extraction (orange), enabling flexible information seeking without rigid chunking. (3) Dense Utility Supervision: providing additional fine-grained feedback by quantifying the utility of each extracted document via relative information gain.}
    \label{fig:2}
\end{figure}

\paragraph{Reward Design in Reinforcement Learning.}
The design of reward functions is a pivotal factor~\citep{ouyang2022training,kwon2023reward,yu2026groundanythingreconcilingparalleldecoding}. Previous methods use reward models in RL and approaches like RewardAnything~\citep{yu2025rewardanything}, GenPRM~\citep{zhao2025genprm}, and others~\citep{guo2025rrm, chen2025rmr1, li2025generalist} leverage the semantic understanding of LLMs to evaluate intermediate reasoning steps or generalizable objectives. Similarly, in the long-context domain, LongReward~\citep{zhang-etal-2025-longreward} utilizes LLM feedback to score multidimensional quality. However, relying on learned RMs presents a challenge, as reward hacking remains difficult to completely mitigate. Consequently, recent breakthroughs such as DeepSeek-R1~\citep{guo2025deepseek} have pivoted back to rule-based, outcome-only rewards, relying on verifiable signals to drive robust self-evolution without hacking risks. Yet, in long-context scenarios, the extended reasoning horizon renders outcome signals excessively sparse~\citep{longrlvr2025}; we mitigate this by introducing additional relative information gain as an intrinsic dense reward, offering better robustness.

\section{Method}

\subsection{Problem Formulation}
\label{sec:preliminary}

\paragraph{RL for Long-Context Reasoning.}
We formulate the long-context question answering task as a Markov Decision Process (MDP). Unlike standard language modeling which conditions solely on a query $x$, our policy $\pi_{\theta}$ must process a lengthy document $c$ to derive the answer $y$. The training objective is to maximize the expected reward while maintaining a trust region constraint relative to a reference model $\pi_{\text{ref}}$:
\begin{equation}
    \max_{\pi_\theta}\; \mathbb{E}_{\substack{x, c \sim \mathcal{D} \\ y \sim \pi_{\theta}(\cdot | x, c)}} \Big[ r_{\phi}(x, c, y)
    - \beta\, \mathbb{D}_{\text{KL}} \big( \pi_{\theta}(\cdot | x, c) \| \pi_{\text{ref}}(\cdot | x, c) \big) \Big]
\end{equation}
where $r_{\phi}$ denotes the reward function that evaluates the correctness and quality of the reasoning chain $y$ given the context $c$.

\subsection{Supervised Fine-tuning (Cold Start)}
\label{sec:sft}

To mitigate the instability of training reinforcement learning from scratch, we employ a warm-up stage using Supervised Fine-Tuning (SFT), effectively treating it as behavioral cloning. In this phase, the policy $\pi_\theta$ is initialized to mimic expert reasoning behaviors by maximizing the likelihood of high-quality demonstration trajectories. The optimization objective is defined as:
\begin{equation}
\mathcal{L}_{\text{SFT}}(\theta) =
  -\mathbb{E}_{(x,c,y) \sim \mathcal{D}_{\text{SFT}}} \left[ \sum_{t=1}^{|y|} \log \pi_\theta(y_t | x, c, y_{<t}) \right]
\label{eq:sft_loss}
\end{equation}
where $c$ represents the long document, $x$ is the query, and $y$ denotes the ground-truth chain-of-thought sequence. This process serves two critical roles: injecting domain knowledge into the model and strictly enforcing the structural requirements (e.g., the \texttt{<think>} encapsulation) necessary for the subsequent group-based exploration.

\subsection{LongR Training Framework}
\label{sec:longrl_framework}

We introduce LongR, a framework that integrates contextual dense rewards with curriculum learning to enhance long-context understanding capabilities.

\subsubsection{Intrinsic Grounding Reward via Information Gain}
\label{sec:reward_modeling}

In long-context reasoning tasks, optimizing policies solely via sparse outcome rewards poses significant optimization challenges. As identified by LongRLVR~\citep{longrlvr2025}, relying exclusively on a final scalar feedback can lead to vanishing gradients for the contextual grounding process. Consequently, the model fails to learn effective contextual grounding, as the supervision signal for identifying relevant evidence vanishes due to the vast search space of the long context.

Conventionally, the reward function relies on a combination of format constraints and final answer correctness. Let $y$ be the generated trajectory and $y^*$ be the ground truth. The standard sparse reward $r_{\text{base}}$ is formulated as:
\begin{equation}
r_{\text{base}}(y, y^*) = r_{\text{fmt}}(y) + r_{\text{acc}}(y, y^*),
\end{equation}
where $r_{\text{fmt}}(\cdot)$ is an indicator function that ensures the output adheres to the required structure (e.g., enclosing reasoning within \texttt{<think>} tags), and $r_{\text{acc}}(\cdot)$ verifies whether the final answer matches $y^*$.

We augment the standard objective with a dense verification signal. We formulate the total reward $r_{\text{total}}$ as:
{
\begin{equation}
r_{\text{total}}(x, c, y) = r_{\text{base}}(y, y^*) + r_{\text{ctx}}(x, c, y),
\end{equation}
}
where $r_{\text{ctx}}(x, c, y)$ denotes contextual reward, which quantifies contextual utility via information gain.

\paragraph{Probabilistic Grounding Verification.}
To facilitate an interleaved process of reasoning and document consultation, we employ a specialized prompt template (see~\cref{sys_template}). This constraint compels the policy $\pi_\theta$ to explicitly distinguish between internal reasoning and external citations. Consequently, based on the generated trajectory $y$, we identify the set of relevant context segments $\mathcal{S} = \{s_1, s_2, \dots, s_K\}$ that the model explicitly consulted. These segments represent the concrete information the model processed to facilitate long-context understanding.

\paragraph{Quantifying Contextual Utility via Mutual Information.}
We measure the contextual utility of each consulted segment $s_k \in \mathcal{S}$. We approach this through an Information-Theoretic lens, positing that if the model genuinely ``reads'' and relies on the document, conditioning on the query $x$ and context $c$ should reduce uncertainty regarding $s_k$.

Formally, we define this utility as the pointwise information gain~\citep{cover1999elements}, calculated as a token-normalized log-ratio between the posterior conditioned on $(x,c)$ and the prior without the query or document:
{
\vspace{-6pt}
\begin{equation}
\text{Score}(s_k,x,c) \approx \frac{1}{T}\log\frac{P(s_k\mid x,c)}{P(s_k)}
=\frac{\log P(s_k\mid x,c)-\log P(s_k)}{T}.
\end{equation}
}
Here $T$ is the number of tokens in the quoted segment. To compute the utility score objectively, we employ a frozen verifier model $\pi_{\text{ver}}$, distinct from the optimizing policy. We utilize Qwen3-4B\footnote{\url{https://huggingface.co/Qwen/Qwen3-4B}} as $\pi_{\text{ver}}$ for two primary reasons: (1) \textbf{Computational efficiency}: using a larger model to score every generated segment during RL training would impose prohibitive computational overhead; (2) \textbf{Representational adequacy}: we assume that a model pre-trained on an extensive corpus which is 36T tokens as reported in~\citep{yang2025qwen3} provides a sufficiently accurate estimate of the true conditional and marginal probabilities for our purpose of quantifying information gain. This choice represents a pragmatic trade-off.

The utility is quantified using token-level negative log-likelihood (NLL) derived from this verifier. For a segment $s_k$ consisting of $T$ tokens $(x_1, \dots, x_T)$, the NLL is defined as:
\begin{equation}
\mathcal{L}(s_k | \text{ctx}) = -\frac{1}{T} \sum_{t=1}^{T} \log \pi_{\text{ver}}(x_t | x_{<t}, \text{ctx}).
\end{equation}

Consequently, the final score is computed as the uncertainty reduction captured by the verifier:
{
\begin{equation}
\text{Score}(s_k,x,c)=\mathcal{L}(s_k)-\mathcal{L}(s_k\mid x,c).
\label{eq:absolute}
\end{equation}
}
The choice of NLL as the underlying metric is that a reduction in NLL serves as a robust proxy for the presence of valid long-range dependencies~\citep{DBLP:conf/iclr/BaiZLZZ0D0L25}. By maximizing this NLL-based mutual information, we explicitly incentivize the model to use context effectively.

\paragraph{Reward Normalization.}
While the raw score captures the absolute information gain, its magnitude is sensitive to the intrinsic complexity of the text segment, which can introduce variance instability during reinforcement learning. To derive a scale-invariant signal, we normalize the score by the model's prior uncertainty. This transforms the metric into the relative information gain~\citep{theil1967economics}, representing the fraction of uncertainty explained by the context:
{
\begin{equation}
\frac{\text{Score}(s_k,x,c)}{\mathcal{L}(s_k)}
=\frac{\mathcal{L}(s_k)-\mathcal{L}(s_k\mid x,c)}{\mathcal{L}(s_k)}
=1-\frac{\mathcal{L}(s_k\mid x,c)}{\mathcal{L}(s_k)}.
\label{eq:derivation}
\end{equation}
}
We obtain the final dense reward:
{
\begin{equation}
r(s_k,x,c)=1-\frac{\mathcal{L}(s_k\mid x,c)}{\mathcal{L}(s_k)}.
\label{eq:relative}
\end{equation}
}
This bounded reward $r(s_k,x,c)$ encourages the policy to maximize the \textit{relative} explanatory power of the retrieved context. For $\mathcal{L}(s_k)>0$, the reward is at most 1 but has no finite lower bound. A lower-clipped variant is discussed in Appendix~\ref{sec:revision_clipping}. Finally, the trajectory-level contextual reward is the average score of all unique quotes:
{
\begin{equation}
r_{\text{ctx}}(x,c,y)=\frac{1}{K}\sum_{k=1}^{K}r(s_k,x,c).
\end{equation}
}
If no quotes are generated ($K=0$), the $r_{\text{ctx}}(x, c, y)$ is set to 0.

\subsubsection{Progressive Context Curriculum}
\label{sec:curriculum}

To eliminate the need for costly, long-context-specific data engineering, we adopt a curriculum learning strategy where the model is progressively exposed to documents of increasing length.

Specifically, we partition the training corpus into \(M\) subsets based on document length. In stage \(m\) (\(0 \le m < M\)), the model is trained exclusively on documents whose length does not exceed \(L_m\), where \(L_m = \min(L_{\text{max}}, L_0 \times 2^m)\). For instance, the initial stage (\(m=0\)) might use documents up to 16K tokens, while subsequent stages gradually increase this limit to 32K, and finally the full target length \(L_{\text{max}}\).

\section{Experiments}

\subsection{Set Up}

We conduct experiments using the \texttt{Qwen3} family at both 4B and 8B scales. Our training consists of two stages: supervised fine-tuning (SFT) followed by reinforcement learning (RL) with DAPO~\citep{yu2025dapo}. We employ a two-stage curriculum learning strategy, progressively increasing context length from 16K to 32K tokens.

We evaluate LongR on three long-context benchmarks: \textbf{LongBench v2}~\citep{bai2025longbench}, \textbf{RULER}~\citep{hsieh2024ruler}, and \textbf{InfiniteBench}~\citep{zhang-etal-2024-bench}. Baselines include: (1) officially released \texttt{Qwen3-Base} models; (2) our reproduced \texttt{Qwen3-SFT} models; and (3) \texttt{Qwen3-SFT w/ DAPO} trained with outcome-only rewards. All evaluations use official scripts with a fixed random seed.
Detailed training configurations, hyperparameters, and dataset specifications are provided in Appendix~\ref{sec:appendix_setup}.

\begin{table}[tbp]
\centering
\resizebox{0.85\linewidth}{!}{
\begin{tabular}{p{5.7cm} | c | c c | c @{\hspace{3pt}} c @{\hspace{3pt}} c}
\toprule
 & & \multicolumn{2}{c|}{\textbf{Difficulty}} & \multicolumn{3}{c}{\textbf{Length}} \\
\cmidrule(r){1-2} \cmidrule(lr){3-4} \cmidrule(l){5-7}
\textbf{Model} & \textbf{Overall} & \textbf{Easy} & \textbf{Hard} & \textbf{Short} & \textbf{Medium} & \textbf{Long} \\
\midrule
\texttt{Qwen3-8B*}   & 32.99 & 40.62 & 28.28 & 36.15 & 29.36 & 34.95 \\
\texttt{Qwen3-8B-Base} & 20.58 & 19.66 & 21.14 & 27.85 & 17.76 & 14.06 \\
\texttt{Qwen3-8B-SFT}  & 26.64 & 28.26 & 25.64 & 31.46 & 24.45 & 22.97 \\
\texttt{Qwen3-8B-SFT w/ DAPO} & 32.84 & 39.91 & 28.48 & 38.06 & 28.75 & 32.29  \\
\texttt{Qwen3-8B-SFT w/ LongR} & \textbf{36.23} & \textbf{41.15} & \textbf{33.20} & \textbf{38.75} & \textbf{32.67} & \textbf{39.12} \\
   \cmidrule(lr){1-1}  \cmidrule(lr){2-2} \cmidrule(lr){3-4} \cmidrule(lr){5-7}
\texttt{Qwen3-4B*}   & 31.46 & 34.70 & 29.46 & 35.28 & 28.23 & \textbf{31.54} \\
\texttt{Qwen3-4B-Base}   & 17.13 & 19.60 & 15.61 & 20.83 & 13.69 & 17.82 \\
\texttt{Qwen3-4B-SFT}  & 25.65 & 21.58 & 28.16 & 32.95 & 22.53 & 19.68 \\
\texttt{Qwen3-4B-SFT w/ DAPO} & 30.41 & 28.32 & 31.69 & 33.33 & 31.25 & 23.84 \\
\texttt{Qwen3-4B-SFT w/ LongR} & \textbf{35.44} & \textbf{35.16} & \textbf{35.61} & \textbf{38.75} & \textbf{37.21} & 26.39 \\
\bottomrule
\end{tabular}
}
\caption{Evaluation results (\%) on LongBench v2. Qwen3-8B* and Qwen3-4B* are officially released models available on HuggingFace. Short, Medium, and Long refer to length ranges of <32k, 32-128k, and >128k, respectively. Qwen3-8B* is evaluated using the official thinking mode, and we have set the random seed to 0 to ensure reproducibility.}
\label{tb:main}
\end{table}

\begin{table}[tbp]
    \centering
    \resizebox{0.95\linewidth}{!}{
        \begin{tabular}{l|cccc|cccc}
        \toprule
        \multirow{2}{*}{ \textbf{Model}} & \multicolumn{4}{c|}{\textbf{RULER-128k}} & \multicolumn{4}{c}{\textbf{RULER-64k}} \\
        \cline{2-9}
        ~ & \textbf{NIAH} & \textbf{VT} & \textbf{QA} & \textbf{Avg} & \textbf{NIAH} & \textbf{VT} & \textbf{QA} & \textbf{Avg} \\
        \midrule
        \texttt{Qwen3-8B-Base} & 51.88 & 47.68 & 17.60 & 39.05 & 61.76 & 28.08 & 39.70 & 43.18 \\
        \texttt{Qwen3-8B-SFT} & \textbf{61.64} & 51.96 & 30.50 & 48.03 & 62.96 & 31.00 & 38.80 & 44.25 \\
        \texttt{Qwen3-8B-SFT w/ DAPO} & 57.64 & 76.56 & 30.40 & 54.87 & 66.13 & 58.52 & 40.40 & 55.02 \\
        \texttt{Qwen3-8B-SFT w/ LongR} & 58.63 & \textbf{80.20} & \textbf{45.60} & \textbf{61.48} & \textbf{69.88} & \textbf{86.64} & \textbf{59.42} & \textbf{71.98} \\
        \midrule
        \texttt{Qwen3-4B-Base} & 46.74 & 45.64 & 14.00 & 35.46 & 60.55 & 56.40 & 19.30 & 45.42 \\
        \texttt{Qwen3-4B-SFT} & 53.02 & \textbf{91.44} & 27.20 & 57.22 & 66.20 & 90.92 & 20.20 & 59.11 \\
        \texttt{Qwen3-4B-SFT w/ DAPO} & 54.57 & 90.48 & 28.70 & 57.92 & 67.74 & 91.96 & 30.90 & 63.53 \\
        \texttt{Qwen3-4B-SFT w/ LongR} & \textbf{56.28} & 90.84 & \textbf{35.00} & \textbf{60.71} & \textbf{71.36} & \textbf{93.32} & \textbf{43.60} & \textbf{69.43} \\
        \bottomrule
        \end{tabular}
    }
    \caption{Evaluation results (\%) on RULER-128K and RULER-64K. NIAH denotes Needle-In-A-Haystack retrieval, VT denotes Variable Tracking, and QA denotes multi-hop question answering. }
    \label{tb:ruler}
\end{table}

\begin{table}[tbp]
\centering
\begin{tabular}{l|cccccc|c}
    \toprule
    \bf Methods & \bf Re.Pa & \bf Re.Nu & \bf En.Sum & \bf En.QA & \bf Zh.QA & \bf En.MC  & \bf Avg. \\
    \midrule
    \texttt{Qwen3-8B-SFT}  & 85.76 & 86.44 & 14.18 & 25.09 & 14.47 & 54.15 & 46.68 \\
    \texttt{Qwen3-8B-SFT w/ DAPO}  & 85.93 & 83.56 & 11.86 & \textbf{27.45} & \textbf{23.54} & 56.33 & 48.11 \\
    \texttt{Qwen3-8B-SFT w/ LongR}  & \textbf{87.46} & \textbf{89.79} & \textbf{15.72} & 27.40 & 22.31 & \textbf{58.52} & \textbf{ 50.20 } \\
    \midrule
    \texttt{Qwen3-4B-SFT}  & 86.78  & 83.43 & 12.76 & 12.68 & 9.65 & 53.28 & 43.10 \\
    \texttt{Qwen3-4B-SFT w/ DAPO}  & \textbf{87.63} & 84.75 & \textbf{17.78} & 14.83  & 10.39 & 54.59 & 44.99  \\
    \texttt{Qwen3-4B-SFT w/ LongR}  & 86.95 & \textbf{87.08} & 16.07 & \textbf{21.85} & \textbf{17.39}  & \textbf{56.33}  & \textbf{ 47.61 }  \\
    \bottomrule
\end{tabular}
\caption{Evaluation results (\%) on InfiniteBench. Re.Pa and Re.Nu denote Retrieve.PassKey and Retrieve.Number, En.Sum denotes English Summarization, En.QA and Zh.QA denote English and Chinese question answering, and En.MC denotes English multiple-choice reasoning. }
\label{tb:infinite}
\end{table}

\begin{table}[tbp]
\centering
\resizebox{0.95\linewidth}{!}{
\begin{tabular}{p{5.7cm} | c | c c | c @{\hspace{3pt}} c @{\hspace{3pt}} c}
\toprule
 & & \multicolumn{2}{c|}{\textbf{Difficulty}} & \multicolumn{3}{c}{\textbf{Length}} \\
\cmidrule(r){1-2} \cmidrule(lr){3-4} \cmidrule(l){5-7}
\textbf{Model} & \textbf{Overall} & \textbf{Easy} & \textbf{Hard} & \textbf{Short} & \textbf{Medium} & \textbf{Long} \\
\midrule

\texttt{Qwen3-8B-SFT} & 26.64 & 28.26 & 25.64 & 31.46 & 24.45 & 22.97 \\
\rowcolor{gray!20} \multicolumn{7}{l}{\textbf{Reinforcement Learning}} \\
\texttt{Outcome-only Reward}  & 32.80 & 40.10 & 28.30 & 38.33 & 28.37 & 32.41 \\
\texttt{w/ Absolute Density}  & 32.41 & \textbf{41.41} & 26.85 & 35.97 & \textbf{33.14} & 25.00 \\
\texttt{w/ Threshold Density} & \underline{34.10}  & 35.16 & \textbf{33.44} & \textbf{41.25} & 26.74 & 36.81 \\
\texttt{w/ Relative Density (Ours)} & \textbf{36.23} & \underline{41.15} & \underline{33.20} & \underline{38.75} & \underline{32.67} & \textbf{39.12} \\

\cmidrule(lr){1-1} \cmidrule(lr){2-2} \cmidrule(lr){3-4} \cmidrule(lr){5-7}

\texttt{Qwen3-4B-SFT}  & 25.65 & 21.58 & 28.16 & 32.95 & 22.53 & 19.68 \\
\rowcolor{gray!20} \multicolumn{7}{l}{\textbf{Reinforcement Learning}} \\
\texttt{Outcome-only Reward}  & 30.42 & 28.12 & 31.83 & 33.33 & 31.16 & 24.07 \\
\texttt{w/ Absolute Density}  & 31.61 & 30.21 & \underline{32.48} & 36.67 & \underline{31.63} & 23.15 \\
\texttt{w/ Threshold Density} & \underline{33.95} & \textbf{38.02} & 31.43 & \textbf{40.14} & 29.77 &  \textbf{31.94} \\
\texttt{w/ Relative Density (Ours)} & \textbf{35.44} & \underline{35.16} & \textbf{35.61} & \underline{38.75} & \textbf{37.21} & \underline{26.39} \\

\bottomrule
\end{tabular}
}
\caption{Ablation study on Reward Design. We compare our proposed method against three baselines: (1) Outcome-only, which relies \textbf{exclusively} on sparse final signals ($r_{acc}$ and $r_{fmt}$) without dense supervision; (2) w/ Absolute Density, which substitutes the relative metric with the absolute information gain (Eq.~\ref{eq:absolute}) as the dense signal; and (3) w/ Threshold Density, which applies a binary reward scheme, assigning a discrete value of 1 only if the relative gain (Eq.~\ref{eq:relative}) exceeds 0.5, and 0 otherwise. Evaluation results (\%) on LongBench v2.}
\label{tb:reward}
\end{table}

\subsection{Main Results}
\label{sec:main_results}

\paragraph{Results on LongBench v2.}
\cref{tb:main} presents results on LongBench v2. LongR yields substantial gains over both the SFT baseline and outcome-only RL (DAPO). For \texttt{Qwen3-8B}, LongR achieves an overall score of 36.23, improving over DAPO by +3.39 and over SFT by +9.59. The \texttt{Qwen3-4B} model shows a similar trend, with LongR reaching 35.44. Overall, outperforming DAPO by +5.03 and SFT by +9.79.

Notably, LongR delivers the strongest improvements on challenging examples: Hard questions improve from 25.64 to 33.20 (8B) and from 28.16 to 35.61 (4B), while the Long split rises from 22.97 to 39.12 (8B). These results indicate that our method enables more effective use of document context over shallow heuristics.

\subsection{Ablation on More Long-context Benchmarks}
\label{sec:ablation_more_benchmark}

\cref{tb:ruler} shows that LongR consistently improves performance on the RULER benchmark at both 128K and 64K context lengths. For the 8B model, LongR achieves average scores of 61.48 (128K) and 71.98 (64K), outperforming the outcome-only DAPO baseline by +6.61 and +16.96 respectively. The 4B model shows a similar trend, with LongR attaining 60.71 (128K) and 69.43 (64K), leading in all settings. Notable gains appear in VT and QA for 8B model, and primarily in QA for 4B model.

\cref{tb:infinite} presents results on InfiniteBench, where LongR again achieves the highest average scores: 50.20 for 8B and 47.61 for 4B. It delivers substantial gains on key subtasks, most notably improving En.MC from 54.15 to 58.52 (8B) and boosting En.QA from 12.68 to 21.85 (4B).

\subsection{Impact of Reward Design Choices}

To validate the effectiveness of our proposed reward design, we conduct an ablation study comparing our method against three distinct reward configurations.
As presented in~\cref{tb:reward}, Relative Infor- mation Gain yields the superior performance. The results reveal critical insights into reward design:

\textbf{Outcome-only Reward.} The sparse outcome signal is insufficient to guide the policy through the vast search space of long-context reasoning.

\textbf{w/ Absolute Density.} This variant underperforms because the raw information gain is unbounded and sensitive to intrinsic text complexity. The policy tends to hack the reward by generating unnecessarily verbose or esoteric segments to maximize prior uncertainty ($\mathcal{L}(s_k)$), rather than focusing on true relevance.

\textbf{w/ Threshold Density.} %
It indiscriminately rewards all segments above the cutoff equally, failing to distinguish between moderately relevant text and precise evidence.

\textbf{w/ Relative Density (Ours).} Our method effectively acts as a scale-invariant filter. For irrelevant or parametric content, the context offers no new predictive power, resulting in similar prior and posterior likelihoods ($\mathcal{L}(s_k|c) \approx \mathcal{L}(s_k)$), which drives the reward to near zero. Conversely, for specific, context-dependent evidence (e.g., unique entities or dates), the document significantly reduces the high initial uncertainty ($\mathcal{L}(s_k|c) \ll \mathcal{L}(s_k)$), resulting in a high reward. This mechanism naturally encourages the model to extract information that is maximally informative yet grounded, regardless of the segment's length.
A detailed case study in~\cref{fig:case} illustrates how our method works.

\begin{table}[tbp]
\centering
\resizebox{0.85\linewidth}{!}{
\begin{tabular}{p{5.7cm} | c | c c | c @{\hspace{3pt}} c @{\hspace{3pt}} c}
\toprule
 & & \multicolumn{2}{c|}{\textbf{Difficulty}} & \multicolumn{3}{c}{\textbf{Length}} \\
\cmidrule(r){1-2} \cmidrule(lr){3-4} \cmidrule(l){5-7}
\textbf{Model} & \textbf{Overall} & \textbf{Easy} & \textbf{Hard} & \textbf{Short} & \textbf{Medium} & \textbf{Long} \\
\midrule

\texttt{Qwen3-8B-SFT} & 26.64 & 28.26 & 25.64 & 31.46 & 24.45 & 22.97 \\
\hspace{1.0em}\texttt{+LongR w/ 1.7B verifier}  & 34.59 & 39.06 & 31.83 & 43.33 & 26.98 & 35.19 \\
\hspace{1.0em}\texttt{+LongR w/ 4B verifier} & 36.23  & 41.15 & 33.20 & 38.75 & 32.67 & 39.12 \\
\hspace{1.0em}\texttt{+LongR w/ 32B verifier} & 36.83 & 41.67 & 33.84 & 38.75 & 33.14 & 40.97 \\

\cmidrule(lr){1-1} \cmidrule(lr){2-2} \cmidrule(lr){3-4} \cmidrule(lr){5-7}

\texttt{Qwen3-4B-SFT}  & 25.65 & 21.58 & 28.16 & 32.95 & 22.53 & 19.68 \\
\hspace{1.0em}\texttt{+LongR w/ 1.7B verifier}  & 35.09 & 32.03 & 36.98 & 37.08 & 33.37 & 35.19 \\
\hspace{1.0em}\texttt{+LongR w/ 4B verifier} & 35.44 & 35.16 & 35.61 & 38.75 & 37.21 & 26.39 \\
\hspace{1.0em}\texttt{+LongR w/ 32B verifier} & 36.08 & 36.20 & 36.01 & 38.61 & 36.40 & 31.25 \\

\bottomrule
\end{tabular}
}
\caption{Ablation study on Verifier Size. We evaluate LongR performance using varying verifier sizes (1.7B, 4B, and 32B) from the \texttt{Qwen3} family. For a controlled comparison, all other components remain fixed.}
\label{tb:verifier}
\end{table}

\begin{table}[tbp]
\centering
\resizebox{0.85\linewidth}{!}{
\begin{tabular}{p{5.7cm} | c | c c | c @{\hspace{3pt}} c @{\hspace{3pt}} c}
\toprule
 & & \multicolumn{2}{c|}{\textbf{Difficulty}} & \multicolumn{3}{c}{\textbf{Length}} \\
\cmidrule(r){1-2} \cmidrule(lr){3-4} \cmidrule(l){5-7}
\textbf{Model} & \textbf{Overall} & \textbf{Easy} & \textbf{Hard} & \textbf{Short} & \textbf{Medium} & \textbf{Long} \\
\midrule

\texttt{Qwen3-8B-SFT} & 26.64 & 28.26 & 25.64 & 31.46 & 24.45 & 22.97 \\
\hspace{1.0em}\texttt{w/ DAPO}  & \textbf{36.23} & \textbf{41.15} & 33.20 & 38.75 & 32.67 & \textbf{39.12} \\
\hspace{1.0em}\texttt{w/ CISPO} & 34.99 & 35.94 & 34.41 & 34.41 & \textbf{36.57} & 32.81 \\
\hspace{1.0em}\texttt{w/ GSPO}  & 35.39 & 33.92 & \textbf{36.29} & \textbf{39.17} & 35.12 & 29.63 \\

\cmidrule(lr){1-1} \cmidrule(lr){2-2} \cmidrule(lr){3-4} \cmidrule(lr){5-7}

\texttt{Qwen3-4B-SFT} & 25.65 & 21.58 & 28.16 & 32.95 & 22.53 & 19.68 \\
\hspace{1.0em}\texttt{w/ DAPO}  & \textbf{35.44} & \textbf{35.16} & 35.61 & 38.75 & \textbf{37.21} & 26.39 \\
\hspace{1.0em}\texttt{w/ CISPO} & 33.95 & 31.51 & 35.45 & \textbf{40.14} & 31.98 & \textbf{27.55} \\
\hspace{1.0em}\texttt{w/ GSPO}  & 33.90 & 30.60 & \textbf{35.93} & 39.03 & 32.91 & 27.31 \\

\bottomrule
\end{tabular}
}
\caption{Ablation study on different RL algorithms. Best results within each backbone are highlighted in \textbf{bold}. Evaluation results (\%) on LongBench v2.}
\label{tb:algo}
\end{table}

\subsection{Compare with Different Sizes of Verifier}
\label{sec:ablation_verifier}

Our contextual reward uses a frozen verifier \(\pi_{\text{ver}}\) to estimate information gain (Eqs.~\ref{eq:absolute}--\ref{eq:relative}). We thus study how verifier scale affects performance to find an efficient balance.

Results in~\cref{tb:verifier} show that while larger verifiers generally improve performance, gains saturate quickly. Moving from a 1.7B to 4B verifier yields clear improvements for both Qwen3-8B (Overall: 34.59$\rightarrow$36.23) and Qwen3-4B (Overall: 35.09$\rightarrow$35.44), particularly on harder and longer tasks. Further scaling to 32B brings only marginal gains (8B: 36.23$\rightarrow$36.83; 4B: 35.44$\rightarrow$36.08), with uneven improvements across task splits. A detailed analysis of training efficiency is presented in~\cref{fig:8}.
These results indicate that a 4B verifier provides an optimal balance: it captures most performance gains while remaining computationally efficient for training-time scoring. We therefore adopt Qwen3-4B as our default verifier.

\subsection{Generalizable Improvements Across RL Algorithms}

We further benchmark two additional variants, \texttt{CISPO} and \texttt{GSPO}, under the same training recipe. Both methods deliver substantial improvements over the \textit{SFT} baselines on LongBench v2 for both 8B and 4B (e.g., +8.35 and +8.75 overall for 8B and +8.30 and +8.25 overall for 4B), achieving a performance comparable to \texttt{DAPO}.
Notably, \texttt{CISPO} and \texttt{GSPO} distribute their gains differently across evaluation splits.
\texttt{CISPO} shows a clear advantage on medium-length instances (36.57\% on the 8B Medium split, the best among all methods), while \texttt{GSPO} is particularly strong on more challenging regimes, achieving the best 8B performance on the Hard split (36.29\%) and the best 4B performance on the Hard split (35.93\%).
These results suggest that our recipe remains effective across different RL algorithms, and that the choice of RL objective mainly affects where the improvements concentrate along the length and difficulty spectrum rather than whether improvements emerge.
Unless otherwise specified, we use \texttt{DAPO} by default due to its strongest and most balanced overall performance.

\section{Conclusion}

We present LongR for enhancing long-context reasoning. LongR introduces a ``Think-and-Read'' pattern that interleaves internal reasoning with explicit document consultation, and proposes a contextual dense reward based on relative information gain to mitigate reward sparsity.

\section*{Limitations}

Due to computational constraints, we were only able to conduct RL training at context lengths up to 32K tokens, as scaling to longer sequences incurs a significantly higher computational budget. We plan to explore scaling to longer contexts in future work.

\FloatBarrier
\bibliographystyle{assets/plainnat}
\bibliography{ref}
\clearpage
\appendix

\begin{figure}[H]
\centering
\begin{minipage}[b]{\ArxivColumnWidth}
{\huge\sffamily Appendix\par}
\vspace{0pt}
\section{Templates}
\label{sec:appendix_templates}
\vspace{-2pt}

    \centering
    \begin{tcolorbox}[
        colframe=blue!75!black,
        colback=blue!5!white,
        coltitle=white,
        title=\textbf{System Prompt Template},
        fontupper=\small
    ]
        You are a rigorous, evidence-driven reasoning assistant. You will be presented with a question based on a provided lengthy document; your task is to answer the question based on the document.

        First, carefully read the query and search for relevant information within the long document. Then, reason based on the quoted evidence to formulate your answer.

        The reasoning process and answer must be enclosed within \texttt{<think>} and \texttt{<answer>} tags, respectively.

        \textit{Example Format:} \\
        \texttt{<think>} reasoning process \texttt{</think>} \texttt{<answer>} answer here \texttt{</answer>}.
    \end{tcolorbox}
    \end{minipage}\hfill
\begin{minipage}[b]{\ArxivColumnWidth}

    \centering
    \begin{tcolorbox}[
        colframe=blue!75!black,
        colback=blue!5!white,
        coltitle=white,
        title=\textbf{System Prompt Template},
        fontupper=\small
    ]
        You are a rigorous, evidence-driven reasoning assistant. You will be presented with a question based on a provided lengthy document; your task is to answer the question based on the document.

        First, carefully read the query and search for relevant information within the long document. When you find a relevant document, you can quote the exact text snippets verbatim within your reasoning. Then, reason based on the quoted evidence to formulate your answer.

        The reasoning process and answer must be enclosed within \texttt{<think>} and \texttt{<answer>} tags, respectively.

        \textit{Example Format:} \\
        \texttt{<think>} reasoning process \dots [Relevant text from the document] \dots \texttt{</think>} \texttt{<answer>} answer here \texttt{</answer>}.
    \end{tcolorbox}
    \end{minipage}
\par
\begin{minipage}[t]{\ArxivColumnWidth}
\caption{The system prompt used for outcome-only RL.}
    \label{baseline_sys_template}
\end{minipage}\hfill
\begin{minipage}[t]{\ArxivColumnWidth}
\caption{The system prompt template used for interleaved reasoning and document consultation. It explicitly instructs the model to generate structured reasoning paths with quoted evidence.}
    \label{sys_template}
\end{minipage}
\end{figure}

\section{Training and Evaluation Details}
\label{sec:appendix_setup}

\paragraph{Training Setting.}
We conduct experiments using the \texttt{Qwen3} family at both 4B and 8B scales. All training processes are executed on 16 NVIDIA H20 141GB GPUs.
\begin{itemize}
    \item \textbf{Supervised Fine-tuning.} We conduct our training using LLaMA-Factory\footnote{\url{https://github.com/hiyouga/LLaMA-Factory}}. For the cold-start phase, we utilize 20k open-source instruction-following samples from AM-DeepSeek-R1-0528-Distilled\footnote{\url{https://huggingface.co/datasets/a-m-team/AM-DeepSeek-R1-0528-Distilled}} and we employ a cosine learning rate scheduler with a peak rate of 2e-5 and a warmup ratio of 0.03. We set the maximum sequence length to 16,384, with sequence packing and training for 900 steps.
    \item \textbf{Reinforcement Learning.} We use verl\footnote{\url{https://github.com/volcengine/verl}} for RL and use DAPO~\citep{yu2025dapo}.

    \textbf{Curriculum Learning:} We adopt a two-stage curriculum strategy, with each stage comprising 100 training steps and a constant learning rate of 1e-6. The context length is limited to 16K in the initial stage ($m=0$) and extended to 32K in the subsequent stage ($m=1$).

    \textbf{Data:} Following MemAgent~\citep{yu2025memagent}, we synthesize 1,000 samples for each length configuration (16K and 32K). To further bolster reasoning capabilities, we incorporate the DocQA-RL-1.6K dataset\footnote{\url{https://huggingface.co/datasets/Tongyi-Zhiwen/DocQA-RL-1.6K}} during the second stage ($m=1$).

    \textbf{Reward Configuration:} The format reward $r_{fmt}$ is set to 1.0 for valid \texttt{<think>...}\allowbreak\texttt{</think>} and \texttt{<answer>}\allowbreak\texttt{...</answer>} structure, and the accuracy reward $r_{acc}$ is set to 2.0 for correct answers. The contextual density reward $r_{ctx}$ is computed using a frozen \texttt{Qwen3-4B} verifier.

    \textbf{Hyperparameters:} We set the learning rate to 1e-6, the batch size 8, and the rollout number to 16. The maximum sequence length is 32,768, and we allow a maximum output length of 4096 tokens and Ulysses sequence parallel 2~\citep{jacobs2023deepspeedulyssesoptimizationsenabling}.
\end{itemize}

\begingroup\hypersetup{citecolor=black,linkcolor=black,urlcolor=black}
\paragraph{Quotation clarifications.}
Following the format in Figure~\ref{sys_template}, we deterministically extract relevant text enclosed in square brackets (\texttt{[...]}) from the generated reasoning.
\endgroup

\paragraph{Evaluation Setting.}
We employ a comprehensive suite of long-context benchmarks to rigorously evaluate model performance:
\begin{itemize}
    \item \textbf{LongBench v2}~\citep{bai2025longbench}: A challenging benchmark focused on realistic long-context deeper understanding. It employs a multiple-choice question format to facilitate rigorous and objective evaluation.

    \item \textbf{RULER}~\citep{hsieh2024ruler}: A benchmark designed to evaluate the effective context window size through diverse synthetic tasks.

    \item \textbf{InfiniteBench}~\citep{zhang-etal-2024-bench}: A dataset targeting diverse long-context capabilities, covering heterogeneous tasks such as retrieval (Re.Pa, Re.Nu), summarization (En.Sum), QA (En.QA, Zh.QA), and multi-choice reasoning (En.MC).
\end{itemize}

\par
\Needspace{240pt}
\begin{wrapfigure}{r}{\ArxivColumnWidth}
\captionsetup{font=footnotesize}

    \centering
    \includegraphics[width=0.95\linewidth]{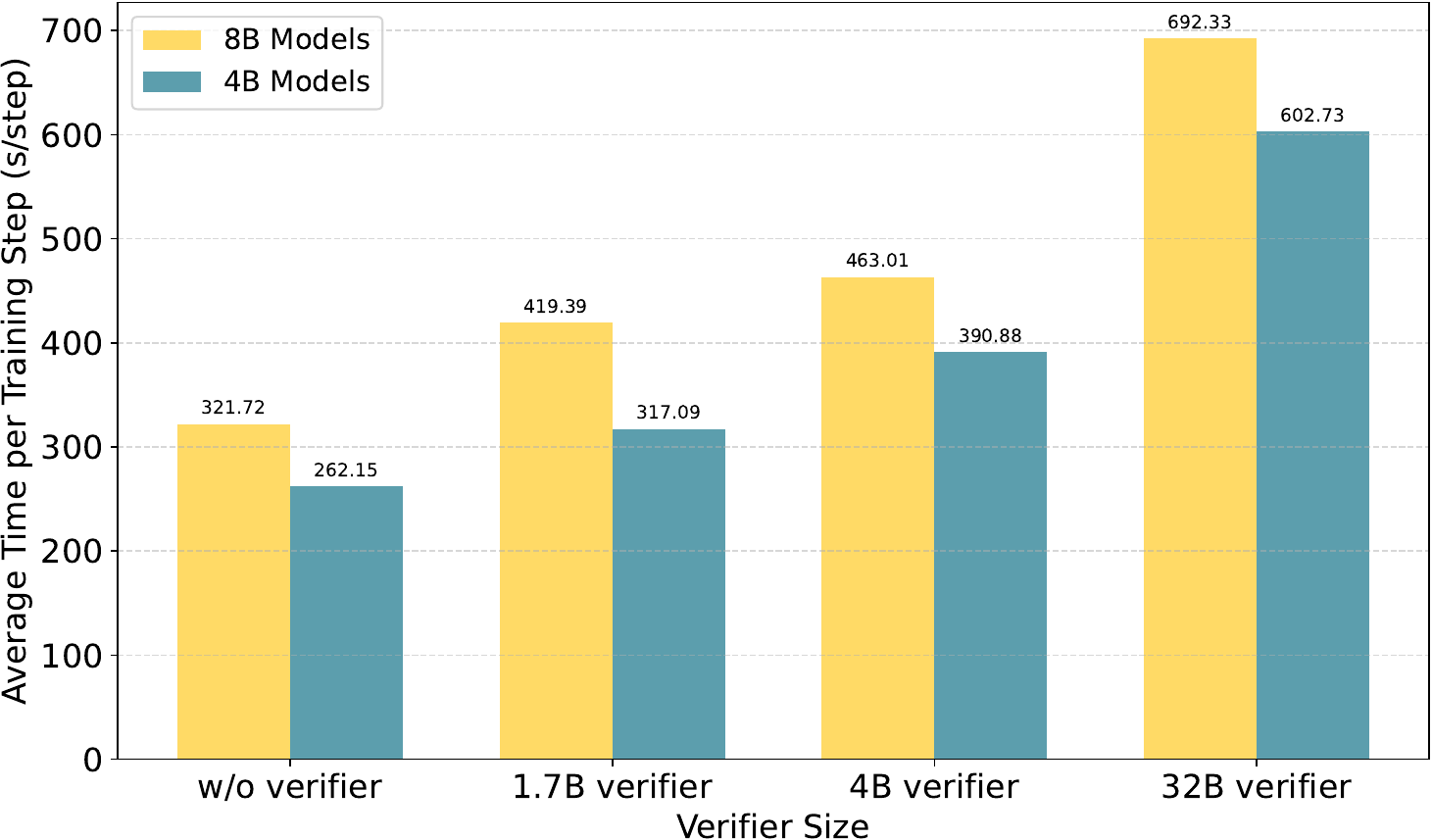}
    \caption{%
 {Average RL training-step time across verifier configurations.} The y-axis ``Average Time per Training Step (s/step)'' %
 {reports the average time per RL training step, in seconds per step.} The average is computed over the first 100 training steps under the experimental setup described in~\cref{sec:appendix_setup}.}
    \label{fig:8}
\end{wrapfigure}

For inference, we utilize vLLM\footnote{\url{https://github.com/vllm-project/vllm}} with a decoding temperature of 1.0. We set maximum generated tokens to 16384. %
 The main LongBench v2 results use decoding seed 0; Appendix~\ref{sec:decoding_seeds} reports evaluation across decoding seeds, and Appendix~\ref{sec:revision_rag} specifies the retrieval baselines' settings.

LongBench v2 results average four runs on the same 503 examples, with the official script's optional 0.25 credit for unparseable answers enabled in every run. We average scores before rounding.

\paragraph{Baselines.}
We conduct comparisons using the \texttt{Qwen3} family at both 4B and 8B scales. The baselines include: (1) the officially released \texttt{Qwen3-Base} models; (2) our reproduced \texttt{Qwen3-SFT} models obtained from the Cold Start phase; and (3) the \texttt{Qwen3-SFT w/ DAPO} variants, which undergo standard full-parameter RL training (Outcome-only) using the same data and steps as LongR with template~\cref{baseline_sys_template} for comparison.

\FinishWrap
\section{Stage-wise Performance of Curriculum Learning}
\label{sec:ablation_curriculum}

\paragraph{Results and analysis.}
As shown in~\mbox{\cref{tb:curriculum}}, curriculum learning yields effective and stable improvements across stages. For \texttt{Qwen3-8B}, the Overall score increases from $26.64$ (\texttt{SFT}) to $32.41$ after the 16K warm-up (Stage $m{=}0$), and further to $36.23$ after extending to 32K (Stage $m{=}1$). The largest gains are observed on the \textbf{Hard} ($33.20$) and \textbf{Long} ($39.12$) subsets, indicating that the model progressively learns to ground reasoning in longer contexts.

The same trend holds for \texttt{Qwen3-4B}: Overall performance rises from $25.65$ to $31.56$ (Stage $m{=}0$) and then to $35.44$ (Stage $m{=}1$), with balanced gains across Difficulty and Length splits. These results validate that progressive context scaling enables the model to first acquire core reasoning skills in shorter contexts and then generalize them effectively to longer documents.

\begin{table}[tbp]
\centering
\resizebox{0.85\linewidth}{!}{
\begin{tabular}{p{5.7cm} | c | c c | c @{\hspace{3pt}} c @{\hspace{3pt}} c}
\toprule
 & & \multicolumn{2}{c|}{\textbf{Difficulty}} & \multicolumn{3}{c}{\textbf{Length}} \\
\cmidrule(r){1-2} \cmidrule(lr){3-4} \cmidrule(l){5-7}
\textbf{Model} & \textbf{Overall} & \textbf{Easy} & \textbf{Hard} & \textbf{Short} & \textbf{Medium} & \textbf{Long} \\
\midrule

\texttt{Qwen3-8B-SFT} & 26.64 & 28.26 & 25.64 & 31.46 & 24.45 & 22.97 \\
\rowcolor{gray!20} \multicolumn{7}{l}{\textbf{W/ Curriculum Learning}} \\
\texttt{{+LongR }w/ Stage m = 0}  & 32.41 & 38.54 & 28.62 & 35.56 & 27.91 & 36.11 \\
\texttt{{+LongR }w/ Stage m = 1}  & 36.23 & 41.15 & 33.20 & 38.75 & 32.67 & 39.12 \\

\cmidrule(lr){1-1} \cmidrule(lr){2-2} \cmidrule(lr){3-4} \cmidrule(lr){5-7}

\texttt{Qwen3-4B-SFT}  & 25.65 & 21.58 & 28.16 & 32.95 & 22.53 & 19.68 \\
\rowcolor{gray!20} \multicolumn{7}{l}{\textbf{W/ Curriculum Learning}} \\
\texttt{{+LongR }w/ Stage m = 0}  & 31.56 & 31.25 & 31.75 & 36.81 & 30.35 & 25.23 \\
\texttt{{+LongR }w/ Stage m = 1}  & 35.44 & 35.16 & 35.61 & 38.75 & 37.21 & 26.39 \\

\bottomrule
\end{tabular}
}
\caption{Detailed contribution of each stage for curriculum learning. {Evaluation results (\%) on LongBench v2.}}
\label{tb:curriculum}
\end{table}

\FloatBarrier
\section{Analysis}

\subsection{Length of Reasoning Chain}

\begin{figure}[!htbp]
\centering
\begin{minipage}[t]{\ArxivColumnWidth}
\captionsetup{font=footnotesize}

    \centering
    \includegraphics[width=0.95\linewidth]{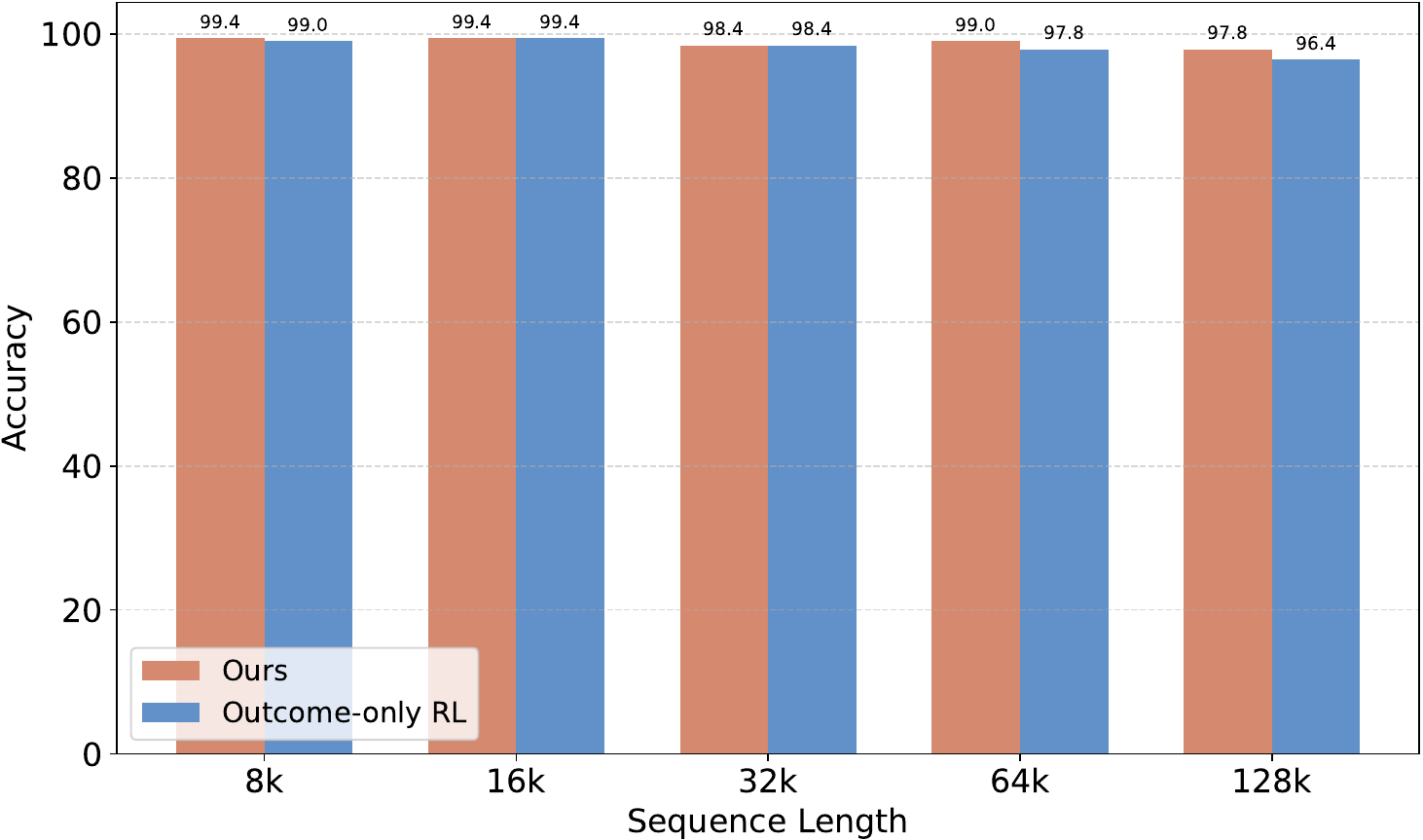}
    \caption{The NIAH accuracy for 8B models.}
    \label{fig:6}
\end{minipage}\hfill
\begin{minipage}[t]{\ArxivColumnWidth}
\captionsetup{font=footnotesize}

    \centering
    \includegraphics[width=0.95\linewidth]{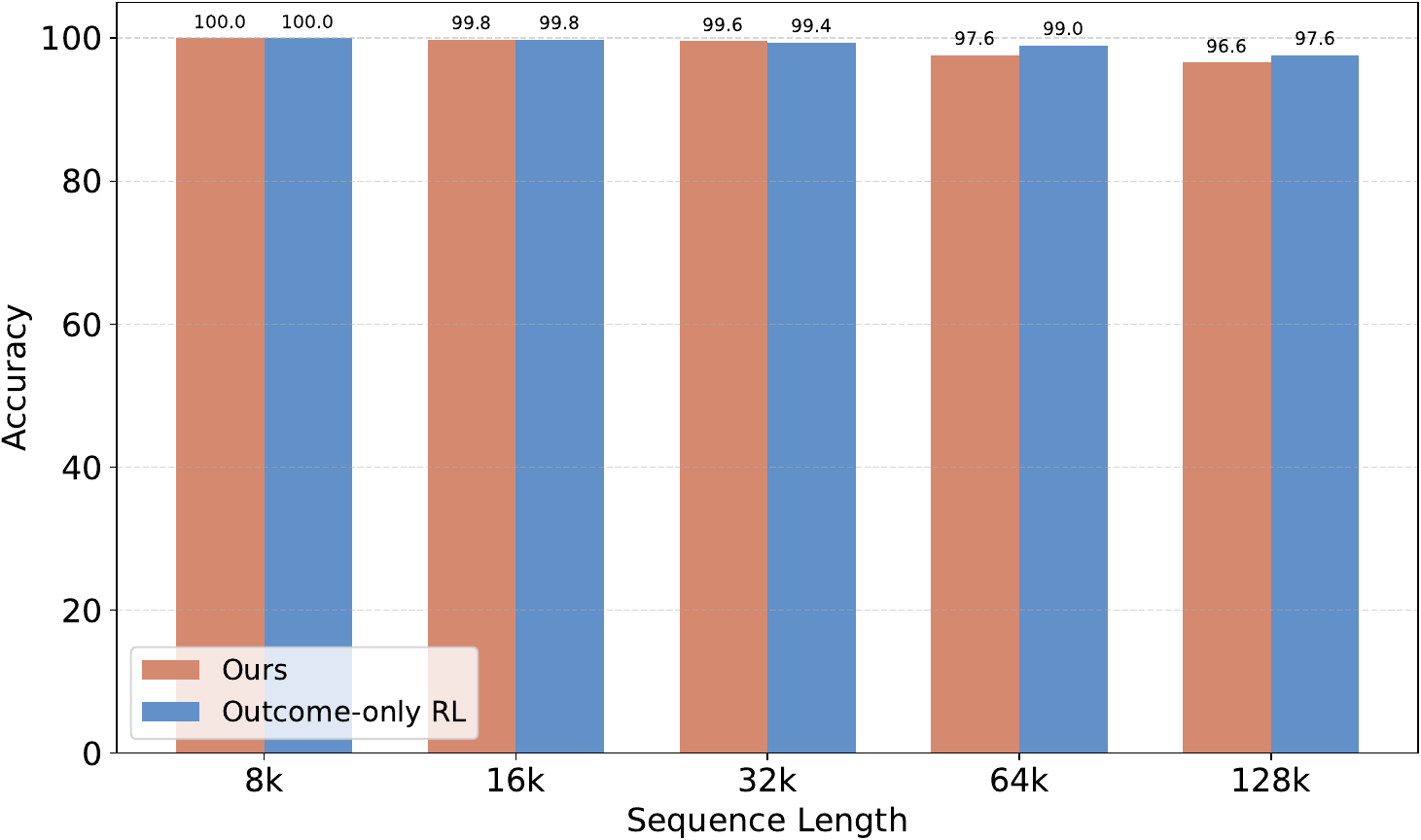}
    \caption{The NIAH accuracy for 4B models.}
    \label{fig:7}
\end{minipage}
\end{figure}
\FloatBarrier
\Needspace{165pt}
\begin{wrapfigure}{r}{\ArxivColumnWidth}
\captionsetup{font=footnotesize}

\captionsetup{font=footnotesize}

    \centering
    \includegraphics[width=1.0\linewidth]{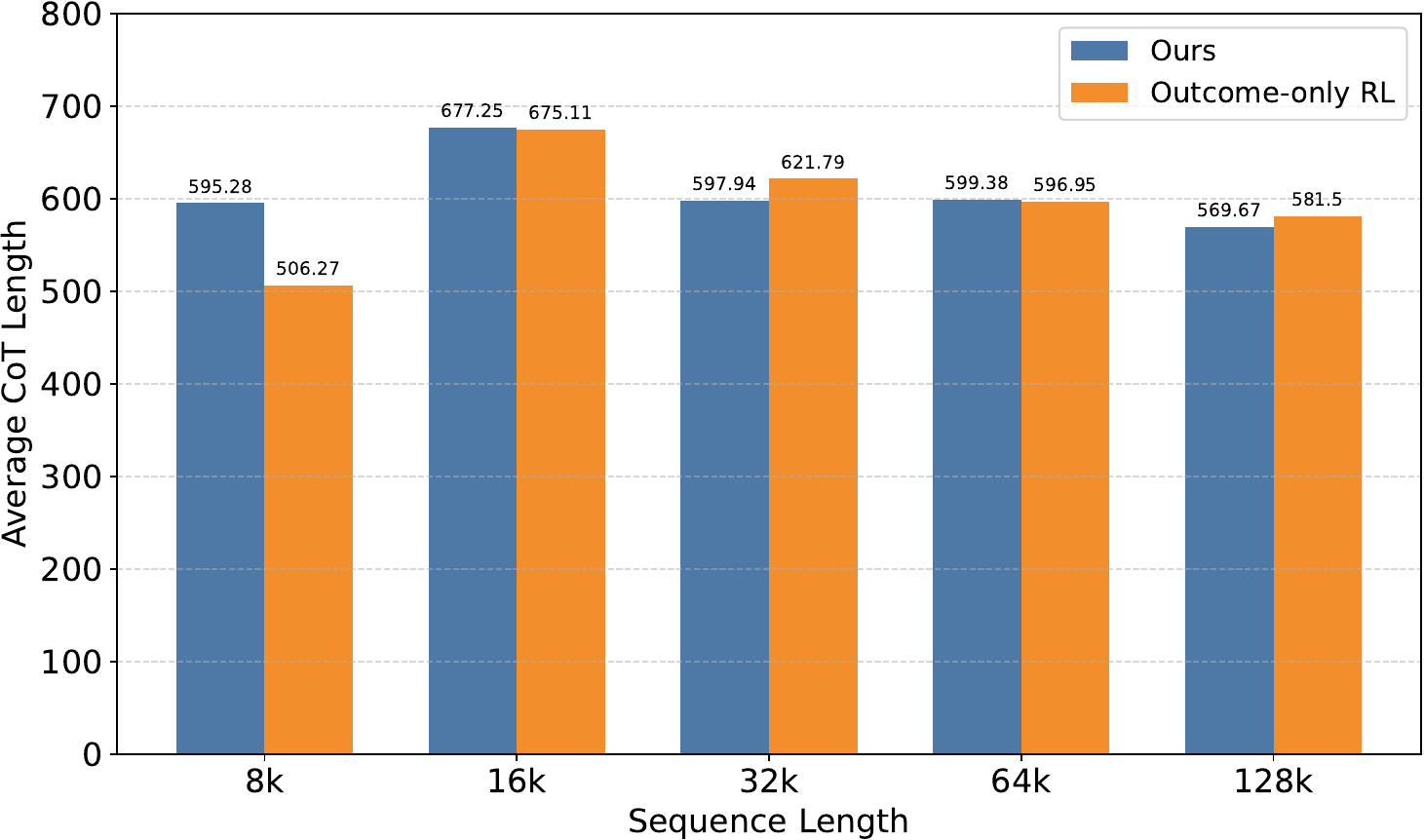}
    \caption{Reasoning chain length comparison for 8B models.}
    \label{fig:3}
\end{wrapfigure}

\begin{wrapfigure}{r}{\ArxivColumnWidth}
\captionsetup{font=footnotesize}

    \centering
    \includegraphics[width=1.0\linewidth]{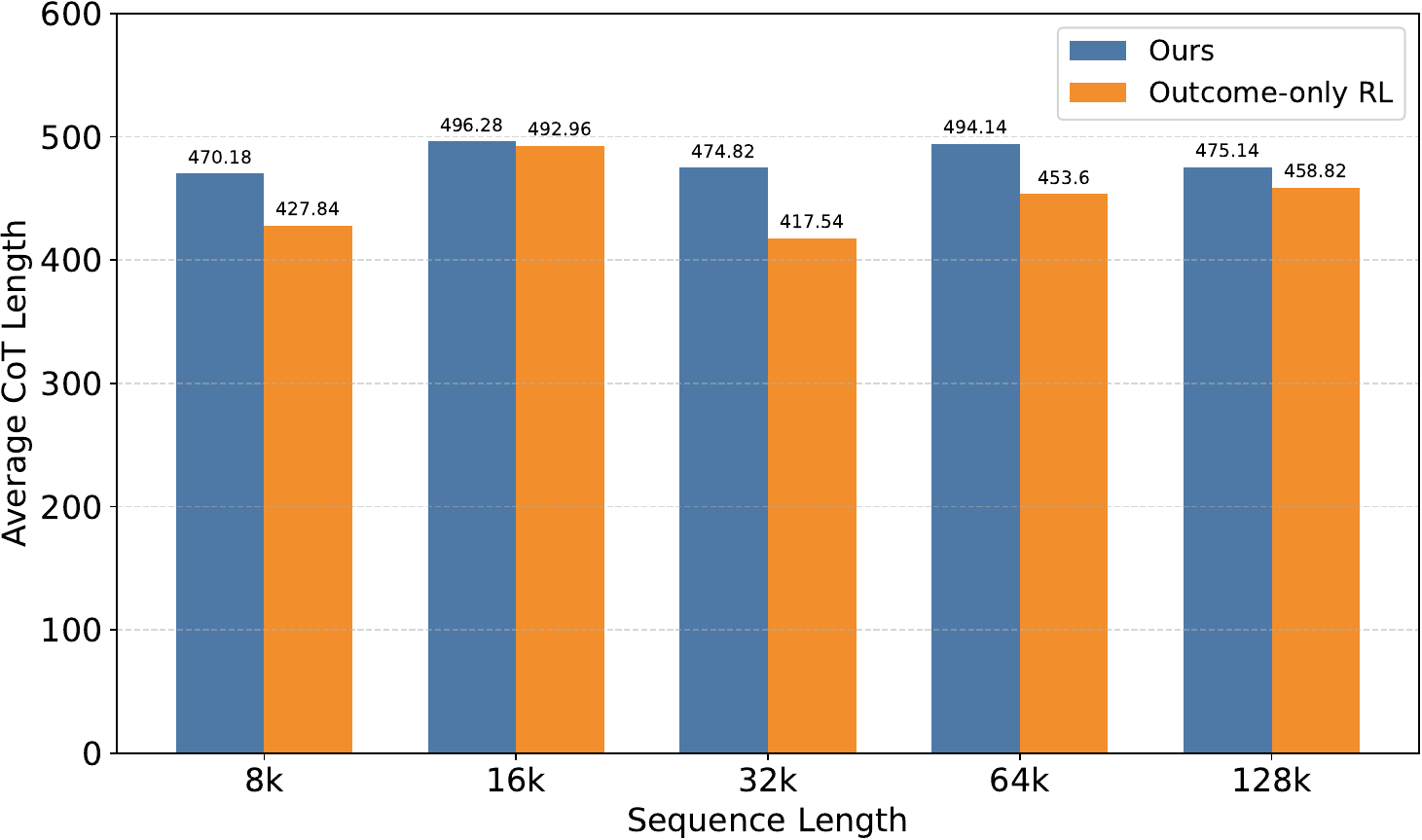}
    \caption{Reasoning chain length comparison for 4B models.}
    \label{fig:4}
\end{wrapfigure}

We investigate whether the introduction of the ``Think-and-Read'' mechanism imposes a significant computational burden or encourages reward hacking behavior (e.g., indiscriminately copying long document segments). Figure~\ref{fig:3} and~\ref{fig:4} illustrate the average length of reasoning chains generated by LongR compared to the Outcome-only baseline on the RULER (NIAH) benchmark across varying context lengths ranging from 8k to 128k. The task's design---embedding a single specific fact within a vast corpus of unrelated text---creates a sharp demarcation between relevant evidence (the needle) and irrelevant noise (the haystack).The corresponding accuracy are shown in~\mbox{\cref{fig:6}} and~\mbox{\cref{fig:7}}.

\FinishWrap
\Needspace{165pt}
\paragraph{Surgical Extraction vs. Length Inflation. }
A primary concern with density-based rewards is that the policy might learn to ``hack'' the metric by blindly copying extensive, irrelevant text chunks to accumulate dense rewards. As observed, LongR maintains an average token count strictly comparable to---and in distinct cases (e.g., 32k, 128k) even marginally lower than---the Outcome-only baseline.

These results support targeted document consultation with comparable reasoning lengths under the evaluated NIAH settings. The query-switching diagnostic in Appendix~\ref{sec:revision_query} complements this behavioral evidence by showing a query-sensitive relative ranking signal for evidence selection in a controlled setting, even when irrelevant quotations receive positive rewards.

\FinishWrap
\Needspace{180pt}
\subsection{Distractors Resistance}

\Needspace{160pt}
\begin{wrapfigure}{r}{\ArxivColumnWidth}
\vspace{-65pt}
\captionsetup{font=footnotesize}

    \centering
    \includegraphics[width=0.95\linewidth]{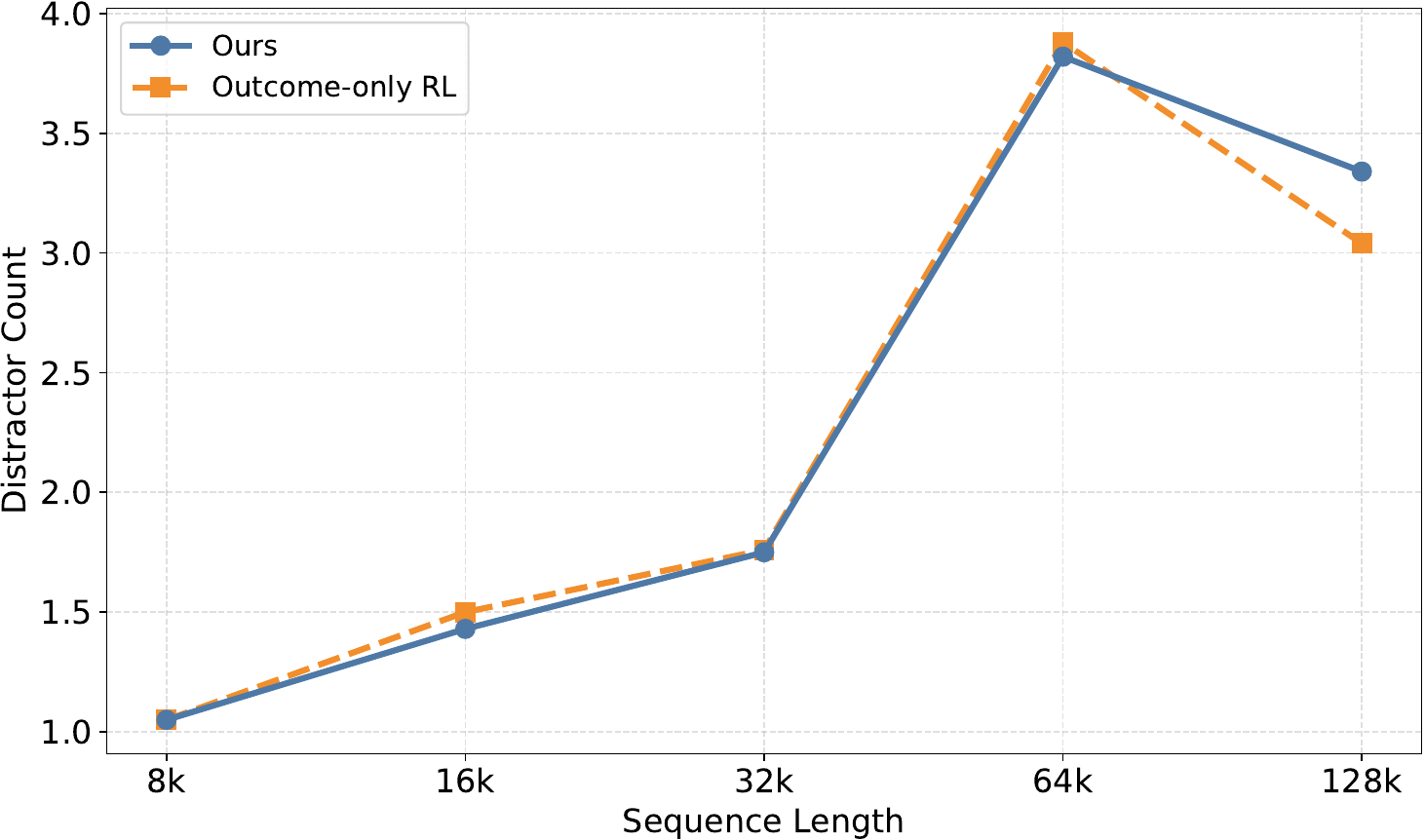}
    \caption{Distractor count for 8B models.}
    \label{fig:5}
\end{wrapfigure}
We further evaluate robustness using the \textbf{Distractor Count} metric on the NIAH task, where any citation other than the specific ``needle'' is considered noise. As shown in Figure~\ref{fig:5}, despite being explicitly incentivized to consult the document via $r_{ctx}$, LongR displays a distractor usage profile nearly identical to the Outcome-only baseline across all context lengths.  This provides behavioral evidence against indiscriminate quotation under the evaluated NIAH settings.

\FinishWrap

\FinishWrap
\FloatBarrier
\begingroup\hypersetup{citecolor=black,linkcolor=black,urlcolor=black}
\FinishWrap
\section{Analysis of the Contextual Reward}
\label{sec:revision_reward}
We analyze the contextual reward along three dimensions: sensitivity to evidence position, reward ranking under query changes, and discrimination between relevant evidence and
rare distractors.

\FinishWrap
\Needspace{180pt}
\subsection{Effect of Evidence Position}

\begin{wraptable}{r}{\ArxivColumnWidth}

\centering\small
\begin{tabular}{lc}
\toprule
Position & Mean reward \\
\midrule
Front (5\%)   & 0.86 \\
Middle (50\%) & 0.86 \\
Tail (95\%)   & 0.85 \\
\bottomrule
\end{tabular}
\caption{{Mean contextual rewards across evidence positions.}}
\label{tab:revision_position}
\end{wraptable}

The verifier may exhibit recency bias, making evidence
near the end of the document easier to predict.
Such evidence could receive a higher contextual reward
because of its position rather than its usefulness.
To examine this possibility, we keep the query,
target evidence, and background text fixed.
We move the complete evidence sentence to sentence
boundaries near 5\%, 50\%, and 95\% of the background
token length.

As shown in Table~\ref{tab:revision_position}, mean rewards remain similar across the three positions, ranging from 0.85 to 0.86. Evidence at the tail receives a slightly lower mean reward than evidence at the front or middle. These results suggest limited variation in mean reward across the tested positions, with no observed tail advantage or middle-position deficit.

\FinishWrap
\Needspace{430pt}
\subsection{Query Sensitivity of the Contextual Reward}
\label{sec:revision_query}

\begin{wrapfigure}{r}{\ArxivColumnWidth}
\captionsetup{font=footnotesize}

    \centering
    \includegraphics[width=0.95\linewidth]{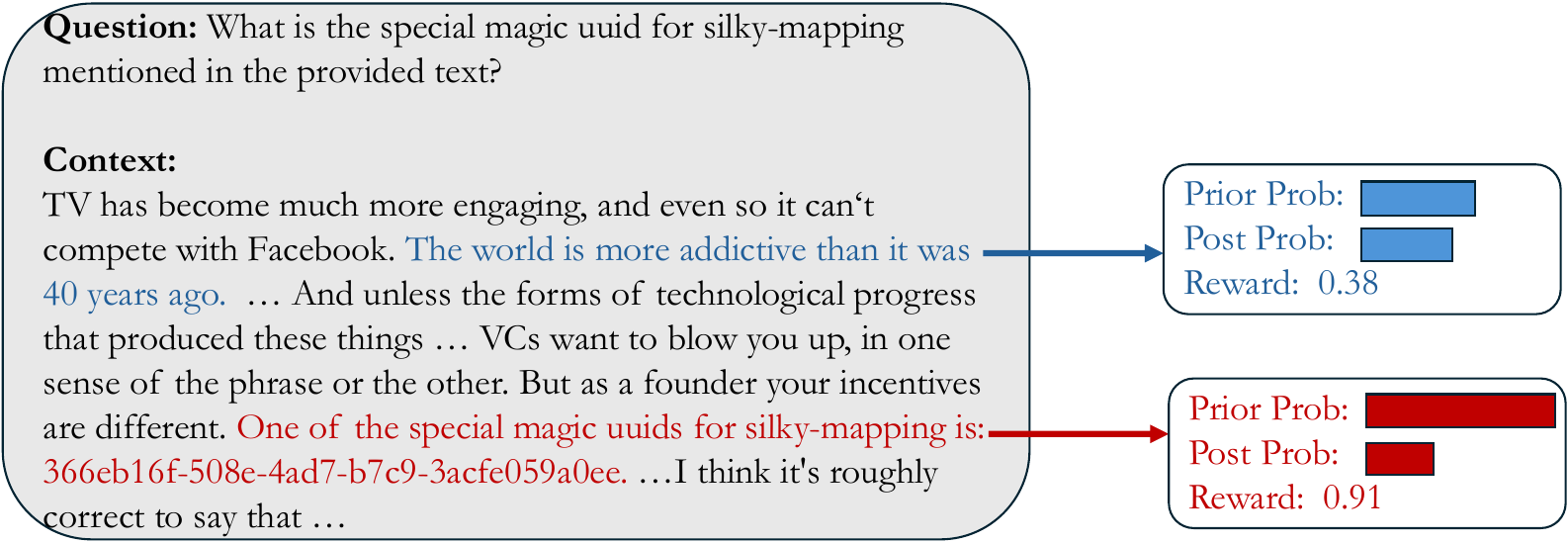}
    \caption{Mechanism Decomposition of Relative Information Gain. We visualize the internal calculation of $r_{ctx}$ on two distinct segments, which are synthesized from RULER. For the Semantic Distractor (blue), the prior loss $\mathcal{L}(s)$ is already low due to the text's high predictability (common sense), resulting in a limited relative gain ($0.38$) despite context consistency. In contrast, the Target Evidence (red) exhibits a high prior loss (unpredictable UUID) which is drastically reduced by the context, yielding a dominant reward ($0.91$). {%
} {This example illustrates a relative reward difference; the two-UUID query-switching diagnostic in Appendix~\ref{sec:revision_query} further tests whether reward preference follows the query.}}
    \label{fig:case}

\captionsetup{type=table,font=footnotesize}

\vspace{10pt}
\centering\small
\setlength{\tabcolsep}{3pt} %
\begin{tabular}{lcc}
\toprule
Query & Expected ranking  & Ranking accuracy \\
\midrule
$x_A$ & $A > B$  & 100.00\% \\
$x_B$ & $B > A$  & 100.00\% \\
\bottomrule
\end{tabular}
\caption{{Reward ranking under query switching.}}
\label{tab:revision_query}
\end{wrapfigure}

A potential concern is that the contextual reward may assign high scores to rare text, even when it is irrelevant to the query. We investigate this concern using randomly sampled UUIDs.
They are difficult to predict without the document,
but providing the document can make them more predictable. Their uniform format also allows us to construct
comparable candidate facts.
We therefore examine whether the reward
can prioritize relevant evidence when both relevant and
irrelevant quotations contain rare UUIDs.

We keep the document and both quotations fixed,
and switch the query to ask about either fact.
If the reward distinguishes query relevance,
the requested fact should receive a higher score
under each query.
We repeat the comparison after swapping the fact positions.

Table~\ref{tab:revision_query} shows all examples achieving
bidirectional success. With the document and quotations
fixed, switching the query reverses the ranking.
This pattern persists after swapping the evidence positions.
 these results support a query-sensitive relative ranking signal for evidence selection in this controlled setting: reward preference is not determined solely by text rarity or copyability. The useful signal is the relevant-minus-irrelevant reward margin and resulting ranking.

\FinishWrap
\Needspace{200pt}
\subsection{Reward Discrimination Across Distractor Types}
\label{sec:revision_rare_distractors}

\begin{wraptable}{r}{\ArxivColumnWidth}

\centering\small
\begin{tabular*}{\linewidth}{@{\extracolsep{\fill}}lrr@{}}
\toprule
Distractor type 
& \shortstack{Mean\\margin} & \shortstack{Ranking\\accuracy} \\
\midrule
Rare person name
  &0.25 & 100.00\% \\
Numerical record
&0.32 & 100.00\% \\
Document-specific fact
& 0.40 & 100.00\% \\
\bottomrule
\end{tabular*}
\caption{{Reward comparison between 
evidence and three types of irrelevant distractors.}}
\label{tab:revision_rare}
\end{wraptable}

Rare names, numerical records, and document-specific
facts may also receive high contextual rewards,
even when they are irrelevant to the query. We introduce irrelevant
sentences containing rare person names, numerical records,
and document-specific facts. These distractors also appear
in the document and may receive high rewards as their
predictability increases. We examine whether query-relevant
evidence still receives higher rewards than these distractors.

For each example, we insert one irrelevant
sentence of each type at a distinct, randomly sampled
sentence boundary.  We score the target evidence
and all three distractors in the same document.

Table~\ref{tab:revision_rare} shows that the target evidence
outranks all three distractor types on all examples.
Mean reward margins are 0.25 for rare person names,
0.32 for numerical records, and 0.40 for document-specific
facts. These results extend the relative ranking observed
in the evidence comparison to the three tested distractor types.

\FinishWrap
\section[Comparison with Retrieval-Augmented Systems]{Comparison with Retrieval-Augmented %
 Systems}
\label{sec:revision_rag}

LongR learns to identify and use evidence within
the input context. An important question is how
it compares with systems that retrieve evidence explicitly.
We evaluate Naive RAG and the public 7B checkpoints
of AutoRefine~\citep{shi2025autorefine} and
R-Search~\citep{zhao2025rsearch} on LongBench v2.

\Needspace{225pt}
\begin{wraptable}{r}{\ArxivColumnWidth}

\centering\footnotesize
\setlength{\tabcolsep}{1pt}
\begin{tabular*}{\linewidth}{@{\extracolsep{\fill}}p{2.2cm}rrrrrr@{}}
\toprule
 & & \multicolumn{2}{c}{\textbf{Difficulty}}
   & \multicolumn{3}{c}{\textbf{Length}} \\
\cmidrule(r){1-2}\cmidrule(lr){3-4}\cmidrule(l){5-7}
\textbf{Model} & \textbf{Overall}
& \textbf{Easy} & \textbf{Hard}
& \textbf{Short} & \textbf{Medium} & \textbf{Long} \\
\midrule

\texttt{R-Search-7B}
& 28.43 & 30.21 & 27.33
& 28.89 & 27.91 & 28.70  \\
\texttt{AutoRefine-7B w/ GRPO}
& 28.23 & 30.21 & 27.01 & 29.44 & 26.05 & 30.56 \\
\texttt{AutoRefine-7B w/ PPO}
& 28.43 & 31.25 & 26.69
& 28.89 & 27.44 & 29.63 \\

\cmidrule(lr){1-1}\cmidrule(lr){2-2}
\cmidrule(lr){3-4}\cmidrule(lr){5-7}

\texttt{Qwen3-8B*}
& 32.99 & 40.62 & 28.28 & 36.15 & 29.36 & 34.95 \\
\texttt{Qwen3-8B* w/ RAG}
& 34.59 & 39.06 & 31.83 & 38.53 & 28.84 & \textbf{39.81} \\
\texttt{Qwen3-8B-SFT w/ LongR}
& \textbf{36.23} & \textbf{41.15} & \textbf{33.20}
& \textbf{38.75} & \textbf{32.67} & 39.12\\

\bottomrule
\end{tabular*}

\caption{{LongBench v2 results (\%).
* denotes public post-trained Qwen3 models evaluated
with non-thinking Naive RAG.}}
\label{tab:revision_rag}
\end{wraptable}

\paragraph{Naive RAG.}
We use a standard retrieve-then-generate pipeline
with the publicly post-trained versions
of Qwen3 at 8B scale.
Evidence is retrieved once using the question
and answer options.

\paragraph{Agentic RAG.}
AutoRefine retains its search and refinement process,
while R-Search retains its search and evidence process.
We evaluate the publicly released AutoRefine-7B-Instruct checkpoint trained with GRPO, along with the PPO- and GRPO-trained variants of R-Search-7B.

\paragraph{Results.}
Table~\ref{tab:revision_rag} shows that LongR outperforms
the evaluated Naive RAG and agentic retrieval baselines
in Overall accuracy. Notably, adding Naive RAG to Qwen3-8B* yields a small Overall
gain, with accuracy slightly increasing from 32.99 to 34.59.This suggests that semantic retrieval alone may not
adequately address the reasoning demands of LongBench v2,
as locally similar passages may omit supporting context
needed to connect evidence and draw conclusions.

Although agentic RAG supports iterative searches,
it remains dependent on query quality and chunk-level
recall, with retrieval errors potentially affecting
subsequent reasoning.
LongR instead integrates evidence consultation directly
into the chain of thought over the provided context,
supporting evidence use as reasoning progresses.

Beyond its stronger reasoning performance in this
evaluation, LongR avoids retrieval-specific overhead.
RAG requires document chunking, embedding, and retrieval,
while agentic RAG additionally involves repeated search
and tool interactions.
LongR operates within standard autoregressive generation
without these separate retrieval operations, simplifying
the inference pipeline and eliminating their associated
computational and orchestration costs.

\FinishWrap
\Needspace{75pt}
\section{Outcome-only Control with a Shared Quotation Prompt}
\label{sec:revision_prompt}

To separate the contribution of the contextual reward
from quotation prompting, we evaluate a Qwen3-8B
outcome-only control.
It uses the same training prompt
(Figure~\ref{sys_template}), inference prompt, and
decoding configuration as LongR.
The control removes the contextual reward while
retaining the format and answer rewards.

\vspace{-10pt}
\Needspace{180pt}
\begin{wraptable}{r}{\ArxivColumnWidth}

\centering\footnotesize\renewcommand{\arraystretch}{0.95}\captionsetup{font=footnotesize}
\setlength{\tabcolsep}{2pt}
\begin{tabular*}{\linewidth}{@{\extracolsep{\fill}}llrr@{}}
\toprule
\multicolumn{2}{l}{Model} & \shortstack{\texttt{Qwen3-8B-SFT}\\\texttt{w/ DAPO}$^\dagger$} & \shortstack{\texttt{Qwen3-8B-SFT}\\\texttt{w/ LongR}} \\
\midrule
 & Overall & 31.81 & \textbf{36.23} \\
\midrule
\multirow{2}{*}{Difficulty} & Easy & 34.90 & \textbf{41.15} \\
 & Hard & 29.90 & \textbf{33.20} \\
\midrule
\multirow{3}{*}{Length} & Short & 33.89 & \textbf{38.75} \\
 & Medium & 31.16 & \textbf{32.67} \\
 & Long & 29.63 & \textbf{39.12} \\
\bottomrule
\end{tabular*}
\caption{LongBench v2 results (\%) with shared training
and inference prompts {and identical decoding settings}.
$^\dagger$ denotes the outcome-only control using
LongR's quotation prompt.
Best scores are bolded.}
\label{tab:revision_prompt}
\end{wraptable}

Table~\ref{tab:revision_prompt} shows that LongR improves
the Overall score from 31.81 to 36.23.
LongR outperforms the control on all five subsets,
with the largest gain on Long,
where the score increases from 29.63 to 39.12.
 With training/inference quotation prompts and decoding settings held fixed, this improvement supports contextual reward as a complement to outcome-only supervision beyond quotation prompting alone on LongBench v2.

\FinishWrap
\Needspace{230pt}
\section{Variability across Decoding Seeds}
\label{sec:decoding_seeds}

\begin{wraptable}{r}{\ArxivColumnWidth}

\centering\footnotesize\renewcommand{\arraystretch}{0.95}\captionsetup{font=footnotesize}
\setlength{\tabcolsep}{2pt}
\begin{tabular*}{\linewidth}{@{\extracolsep{\fill}}llrrrr@{}}
\toprule
\multicolumn{2}{l}{} & \multicolumn{3}{c}{Decoding seed} & \\
\cmidrule(lr){3-5}
& & 0 & 1 & 42 & Mean $\pm$ SD \\
\midrule
 & Overall & 36.23 & 37.77 & 35.79 & $36.60\pm1.04$ \\
\midrule
\multirow{2}{*}{Difficulty} & Easy & 41.15 & 42.71 & 40.62 & $41.49\pm1.08$ \\
 & Hard & 33.20 & 34.73 & 32.80 & $33.57\pm1.02$ \\
\midrule
\multirow{3}{*}{Length} & Short & 38.75 & 40.00 & 38.33 & $39.03\pm0.87$ \\
 & Medium & 32.67 & 34.42 & 32.09 & $33.06\pm1.21$ \\
 & Long & 39.12 & 40.74 & 38.89 & $39.58\pm1.01$ \\
\bottomrule
\end{tabular*}
\caption{LongBench v2 results (\%) for
\texttt{Qwen3-8B-SFT w/ LongR} across three decoding seeds.
SD denotes the sample standard deviation across
decoding runs with a fixed checkpoint. {SD is expressed in percentage points.}}
\label{tab:decoding_seeds}
\end{wraptable}

We evaluate the sensitivity of LongR's performance
to stochastic decoding on LongBench v2.
Using a fixed \texttt{Qwen3-8B-SFT w/ LongR} checkpoint,
we repeat evaluation with decoding seeds 0, 1, and 42,
keeping all other evaluation settings unchanged.
Table~\ref{tab:decoding_seeds} reports the individual
scores and their mean and sample standard deviation
across the three runs.

Table~\ref{tab:decoding_seeds} shows Overall scores
ranging from 35.79 to 37.77, with a mean of 36.60.
 All three scores exceed the shared-prompt outcome-only control's 31.81 (Table~\ref{tab:revision_prompt}); this comparison assesses decoding variability for a fixed LongR checkpoint.

\FinishWrap
\section{Lower-bound Clipping of the Contextual Reward}
\label{sec:revision_clipping}

For positive prior NLL, the contextual reward in
Eq.~\ref{eq:relative} is upper-bounded by 1 but has
no finite lower bound.
To limit extreme negative rewards, we evaluate
a variant with lower-bound clipping:
\begin{equation}
r_{\mathrm{clip}}(s_k,x,c)
= \max\{-1,\,r(s_k,x,c)\}.
\end{equation}

\begin{wraptable}{r}{\ArxivColumnWidth}

\centering\footnotesize\renewcommand{\arraystretch}{0.95}\captionsetup{font=footnotesize}
\setlength{\tabcolsep}{2pt}
\begin{tabular*}{\linewidth}{@{\extracolsep{\fill}}llrr@{}}
\toprule
\multicolumn{2}{l}{Model} & \shortstack{\texttt{Qwen3-8B-SFT}\\\texttt{w/ LongR}} & \shortstack{\texttt{Qwen3-8B-SFT}\\\texttt{w/ LongR}$^\dagger$} \\
\midrule
 & Overall & 36.23 & \textbf{36.98} \\
\midrule
\multirow{2}{*}{Difficulty} & Easy & 41.15 & \textbf{42.19} \\
 & Hard & 33.20 & \textbf{33.76} \\
\midrule
\multirow{3}{*}{Length} & Short & 38.75 & \textbf{39.44} \\
 & Medium & 32.67 & \textbf{33.49} \\
 & Long & 39.12 & \textbf{39.81} \\
\bottomrule
\end{tabular*}
\caption{{LongBench v2 results (\%)
with and without contextual reward clipping.
$^\dagger$ denotes lower-bound clipping at $-1$.
Best scores are bolded.}}
\label{tab:clipping_results}
\end{wraptable}

This restricts the reward to $[-1,1]$ when the original
ratio is defined, while leaving values at or above
$-1$ unchanged.
Clipping does not resolve a zero denominator.
We compare this variant with the original,
unclipped reward on \texttt{Qwen3-8B-SFT w/ LongR}.

Table~\ref{tab:clipping_results} shows that clipping
increases the Overall score from 36.23 to 36.98.
Scores also improve across all five subsets,
with gains ranging from 0.56 to 1.04.
These results suggest that limiting extreme negative
rewards preserves task performance, with modest gains
in this evaluation.

\FinishWrap
\endgroup

\end{document}